\documentclass{article}

\usepackage{arxiv}

\usepackage[utf8]{inputenc} 
\usepackage[T1]{fontenc}    
\usepackage{hyperref}       
\usepackage{url}            
\usepackage{booktabs}       
\usepackage{amsfonts}       
\usepackage{nicefrac}       
\usepackage{microtype}      
\usepackage{lipsum}		
\usepackage{graphicx}
\usepackage{natbib}
\usepackage{doi}

\usepackage[short]{optidef}

\DeclareMathOperator*{\argmin}{arg\,min}

\usepackage{algorithm}
\usepackage[noend]{algpseudocode}

\usepackage{booktabs, multirow}

\usepackage[english]{babel}
\usepackage[autostyle]{csquotes}

\usepackage{graphicx,wrapfig,lipsum}

\setcitestyle{numbers,square,aysep={},yysep={;}}

\title{A New Constructive Heuristic driven by Machine Learning for the Traveling Salesman Problem}

\author{ 
	\href{https://orcid.org/0000-0002-8464-1889}{\includegraphics[scale=0.06]{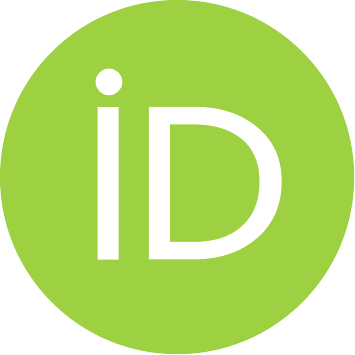}\hspace{1mm}Umberto J.~Mele} \\
	Dalle Molle Institute for Artificial Intelligence\\
	Universit\`{a} della Svizzera Italiana\\
    Lugano, Switzerland\\
	\texttt{umbertojunior.mele@idsia.ch} \\
	\AND
    Luca M.~Gambardella \\
    Dalle Molle Institute for Artificial Intelligence\\
	Universit\`{a} della Svizzera Italiana\\
    Lugano, Switzerland\\
    \And 
    Roberto~Montemanni \\
    Department of Sciences and Methods for Engineering\\
    University of Modena and Reggio Emilia\\
    Reggio Emilia, Italy\\
    \texttt{roberto.montemanni@unimore.it}\\
}

\renewcommand{\shorttitle}{A New Constructive Heuristic driven by ML for the TSP}

\hypersetup{
pdftitle={A New Constructive Heuristic driven by Machine Learning for the Traveling Salesman Problem},
pdfsubject={cs.AI, cs.DM, stat.ML},
pdfauthor={Tommaso~Vitali, Umberto J.~Mele, Luca M.~Gambardella, Roberto~Montemanni},
pdfkeywords={Traveling Salesman Problem, Machine Learning, Artificial Intelligence, Constructive Heuristic, Hybrid Heuristic, Reinforcement Learning, Statistical Analysis, Complexity Theory.},
}

\begin{document}
\maketitle

\begin{abstract}
Recent systems applying Machine Learning (ML) to solve the Traveling Salesman Problem (TSP) exhibit issues when they try to scale up to real case scenarios with several hundred vertices.
The use of Candidate Lists (CLs) has been brought up to cope with the issues.
An initialisation procedure selects a smaller set of promising edges linked to a given vertex, the CL.
These edges are believable to be found in the optimal tour.
The procedure is repeated for each vertex in the TSP, and it allows to restrict the search space during solution creation, consequently reducing the solver computational burden.
So far, ML were engaged to create CLs and values on the edges of these CLs expressing ML preferences at solution insertion.
Although promising, these systems do not clearly restrict what the ML learns and does to create solutions, bringing with them some generalization issues.
Therefore, motivated by exploratory and statistical studies, in this work we instead use a machine learning model to confirm the addition in the solution just for high probable edges.
The restricted use of ML to a simpler and smaller task is motivated by the choice of avoiding the troubles that ML models can encounter with outliers and the detection of under-represented events.
CLs of the high probable edge are employed as input, and the ML is in charge of distinguishing cases where such edges are in the optimal solution from those where they are not. . 
This strategy enables a better generalization and creates an efficient balance between machine learning and searching techniques.
Our \textit{ML-Constructive} heuristic is trained on small instances. 
Then, it is able to produce solutions, without losing quality, to large problems as well.
We compare our results with classic constructive heuristics, showing good performances for TSPLIB instances up to 1748 cities.
Although our heuristic exhibits an expensive constant time operation, we proved that the computational complexity in worst-case scenario, for the solution construction after training, is $O(n^2 \log n^2)$, being $n$ the number of vertices in the TSP instance.
\end{abstract}



\section{Introduction}
The Traveling Salesman Problem (TSP) is one of the most intensively studied and relevant problems in the Combinatorial Optimization (CO) field \cite{Applegate}.
Partially for its simple definition despite its membership to the NP-complete class, partially due to its huge impact on real applications \cite{Matai10}.
The last seventy years have seen the development of an extensive literature, which brought valuable enhancement in the CO branch of study. 
Concepts as the Held-Karp algorithm \cite{Held-Karp}, powerful meta-heuristics as the Ant Colony Optimization \cite{Dorigo}, and effective implementations of local search heuristics as the Lin-Kernighan-Helsgaun \cite{Helsgaun} have been suggested to solve the TSP.
These contributions along with others have supported the development of various applied research domains such as logistics \cite{dellamico}, genetics \cite{Caserta}, telecommunications \cite{telecomunication} and neuroscience \cite{neuroscience}.

In particular, during the last five years, an increasing number of Machine Learning-driven heuristics have appeared to make their contribution to the field \cite{Mele2, Bengio}.
Probably the surge of interest was moved by the rich literature, and by the interesting opportunities that the CO field introduce with its applications. 
Some empowering features which ML models could bring to the field are: the opportunity to leverage knowledge from past solutions \cite{Kool, Mele1},
the ability to imitate computationally expensive operations \cite{Costa}, and the faculty of devising innovative strategies via reinforcement learning paradigms \cite{Zheng}.

In light of the new features being brought by ML approaches, we wish to couple some powerful ML qualities with well known heuristic concepts within a hybrid algorithm,
which seeks robust enhancements with respect to classic approaches. 
The scope is to create an efficient interlocking between ML and optimization heuristics, that strengthens the weaknesses of each single approach.
Many attempts have been proposed so far, but none of them until now has succeeded to preserve the improvements by scaling up to larger problems.
A promising idea to contrast such issue consists in using Candidate Lists (CL) to identify subset of edges \cite{Fu}.
Such subsets help the solver to restrict its searching space during the solution creation,
leading to an incentive for better generalization.

It can be argued that the generalization issue, emerged in the previous ML driven approaches, is caused mostly by the lack of a proper consideration of the ML weaknesses and limitations \cite{DL_limitations, Marcus}.
In fact, ML is known for having troubles with imbalanced datasets, outliers and extrapolation tasks. 
Such events could lead to significant obstacles in achieving good performances with any ML system.
More details on these typical ML weaknesses, with our proposed solutions, will be provided in Section \ref{subsec:2.2}.

Our main contribution is the introduction of the first hybrid system that actively uses ML models to construct partial TSP solutions without losing quality when scaling.
Our \textit{ML-Constructive} heuristic is composed by two phases.
The first uses ML to identify edges that are very likely to be optimal, the second completes the solution with a classic heuristic. 
The resulting heuristic shows good performance, when it is tested on 54 representative instances selected from the TSPLIB library \cite{reinelt1991tsplib}. 
The instances considered present up to 1748 vertices, and surprisingly our \textit{ML-Constructive} exhibits slightly better solutions on larger instances rather than on smaller ones, as shown in the experiments.
Despite good results are shown in terms of quality, our heuristic presents an unappealing large constant time operation in the current state of the implementation.
However, we prove that for the creation of a solution a number of operations bounded by $O(n^2 \log n^2)$ is required after training.
Our ML model learns exclusively using local information, and it employs a simple ResNet architecture \cite{resnet} to recognize some patterns from the candidate lists through images.
The use of images, even if not optimal in term of computation, allowed us to plainly see the input of the network and to get a better understanding of the internal processes in the network.
We have finally introduced a novel loss function to help the network to understand how to make a substantial contribution when building tours.

The TSP is formally stated in Section \ref{subsec:1.1}, and literature review is presented in Section \ref{subsec:1.2}.
The concept of constructive heuristic is described in details in Section \ref{subsec:2.1},
while statistical and exploratory studies on the candidate lists are spotlighted in Section \ref{subsec:2.2}.
The statistical studies were useful to reach the fundamental insights which allow our \textit{ML-Constructive} to bypass the ML weaknesses.
Finally, the general idea of the new method is discussed in Section \ref{subsec:2.3}, the \textit{ML-Constructive} heuristic is explained in Section \ref{subsec:2.4}, and the ML model with the training procedure is discussed in Section \ref{subsec:2.5}.
To conclude, experiments are presented in Section \ref{sec:4}, and conclusions are stated in Section \ref{sec:5}.


\subsection{The Traveling Salesman Problem} \label{subsec:1.1}

Given a graph $G(V, E)$ with $n$ vertices belonging to the set $V=\{0,\dots, n-1\}$,
and pairwise edges $e_{ij} \in E$ for each vertex $i, j \in V$ with $i \neq j$, 
let $c_{ij}$ be the cost for the directed edge $e_{ij}$ starting from vertex $i$ and reaching vertex $j$.
The objective of the Travelling Salesman Problem is to find the shortest possible route 
that visits each vertex in $V$ exactly once, and creates a loop returning to the starting vertex \cite{Applegate}. 

The \citet{dantzig1954solution} formulation for the TSP is an integer linear program describing the requirements that must be met to find the optimal solution for the problem.
The variable $x_{ij}$ defines if the optimal route found picks the edge that goes from vertex $i$ to vertex $j$ with $x_{ij} = 1$, if the route does not pick such edge then $x_{ij}=0$.
A solution is defined as a matrix  $X$ with entries $x_{ij}$ and dimension $n \times n$.
The objective function is to minimize the route cost, as shown in Equation \ref{obj}.

\begin{mini!}|s|[1]
  {}{\sum_{i=0}^{n-1} \; \sum_{j = 0, j \neq i}^{n-1} \quad c_{ij} \, x_{ij}}{}{} \label{obj}
  \addConstraint{\sum_{i=0, i\neq j}^{n-1} x_{ij}=1, \qquad \quad}{j=0,\ldots,n-1} \label{b}
  \addConstraint{\sum_{j=0, j\neq i}^{n-1} x_{ij}=1, \qquad \quad}{i=0,\ldots,n-1} \label{c}
  \addConstraint{\sum_{i \in Q}\sum_{j \in Q, j\neq i} x_{ij} \leq |Q| - 1, \;\;}{\forall Q \subset \{0,\ldots,n-1\}, \,  |Q| \geq 2 } \label{d}
  \addConstraint{x_{ij} \in  \{0,1\}, \qquad \quad }{i, j=0,\ldots,n-1, \quad i \neq j}\label{e}
\end{mini!}

\noindent Subject to the following constraints: each vertex is arrived at from exactly one other vertex (Equation \ref{b}), each vertex is a departure to exactly one other vertex (Equation \ref{c}), and no inner-loop between vertices for any proper subset $Q$ is created (Equation \ref{d}).
The constraints in Equation \ref{d} prevent that the solution $X$ is the union of smaller tours, so it leads to an exponential number of constraints.
Finally, each edge in the solution $x_{ij}$ is not fractional (Equation \ref{e}).

The TSP is called euclidean if the vertices are described by coordinates in the euclidean space and the cost $c_{ij}$ is the euclidean distance between vertices.
The euclidean TSP is also a symmetric, and the euclidean distance satisfies the triangle inequality (Equation \ref{eq:2}) for all triplet of edges.

\begin{gather}
\qquad \qquad c_{jh} \leq c_{ji} + c_{ih},  \qquad \forall \, i,j,h \in V \label{eq:2}    
\end{gather}

\noindent A Candidate List (CL) with cardinality $k$ for vertex $i$ is defined as the set of edges $e_{ij}$, with $j \in CL[i]$, 
such that the vertices $j$ are the closest $k$ vertices to vertex $i$.
Therefore, one of these edges is considered likely to belong to an optimal solution.

\subsection{Literature Review} \label{subsec:1.2}
The first constructive heuristic historically introduced for the TSP is the Nearest Neighbor (NN), 
a greedy strategy that repeats the rule \textit{``take the closest available node''}. 
This procedure is very simple, however it is not very efficient. 
In fact, while the NN is choosing the best available vertex, the Multi-Fragment (MF) \cite{Steiglitz, bentley1990experiments} and the Clarke Wright (CW) \cite{clarke_wright} are alternatives to add the most promising edge in the solution. 
The NN growths a single fragment of tour by adding the closest vertex to the fragment extreme considered during the construction.
On the other hand, MF and CW grow, join and give birth to many fragments of the final tour \cite{Johnson}. 
The approaches using many fragments show superior quality performances, and also come up with very low computational costs.

A different way of constructing TSP tours is known as insertion heuristic, such as the Farthest Insertion (FI) \cite{bentley1990experiments}.
This approach deals with expanding a tour generated by the previous iteration $it$ on a subset of vertices $\hat{V}_{it} \subset V$. 
The expansion is carried out by inserting a new vertex, i.e. $j$, in the subset at each new iteration $\hat{V}_{it + 1} = \hat{V}_{it} \cup \{j\}$.
Therefore, to preserve the feasibility of the expanded tour, one edge is removed from the past tour and two edges are added to connect the released vertices to the new vertex $j$.
The removal is done in such a way that the lowest cost for the new tour is achieved.

In case that the insertion policy is Farthest, the new inserted vertex is always the farthest from the existing vertices in the current tour.
Once the last iterations has been reached, a complete feasible tour passing through each vertex in $V$ has been constructed.

Initially, the exploration in the literature produced to introduce ML concepts was about a setup similar to the NN. 
Systems such as pointer networks \cite{pointer} or graph-based transformers \cite{Kool, Deudon, Mele1} were engaged to select the next vertex in the NN iteration. 
These architecture were engaged to predict a value for each available departure edge of the extreme considered on the single fragment approach. 
Then, the best choice (next vertex) according to these values were added to the fragment. 
These systems were applying stochastic sampling as well, so they could produce thousand of solution in parallel using GPU processing.  
Unfortunately, these ML approaches failed to scale, since all vertices were given as input to the networks, creating confusion to it.

To attempt overcoming the scalability problem, several works proposed systems arguably claimed to be able to generalize \cite{dai2017learning, Ma, Fu}.
Results however showed that, the proposed \textit{structure2vec} \cite{dai2017learning} and Graph Pointer Network \cite{Ma} keep their performances close to the Farthest Insertion (FI) \cite{bentley1990experiments} up to 250 vertices instances, then they lose their ability to generalize.
The model called Att-GCN \cite{Fu} is used to pre-process the TSP by constructing candidate lists.
Even if solutions are promising in this case, the ML was not actively used to construct TSP tours.
Furthermore, no comparisons with other CL constructors as: POPMUSIC \cite{Taillard}, Delunay Triangolarization \cite{Lee}, minimum spanning 1-tree \cite{Applegate} and a recent CL constructor driven by ML \cite{fitzpatrick2021learning} were provided in the paper.

The use of Reinforcement Learning (RL) to tackle the TSP was proposed as well \cite{bello2016neural}.
Where, the ability of learning heuristics without the use of optimal solutions was introduced.
These architectures were trained just via the supervision of the objective function shown in Equation (1a).
The actor-critic algorithm \cite{konda2000actor} was employed as well.
Later, \citet{Mele1} used the actor-critic paradigm with an additional reward defined by the easy to compute minimum spanning tree cost (resembling a TSP solution, see \cite{christofides1976worst}).
Differently, the Q-learning RL framework was used by \citet{dai2017learning}, the Monte Carlo RL was used by \citet{Fu}, and Hierarchical RL introduced in \citet{Ma}.
Finally, \citet{Bresson} explicitly advised to rethink about the generalization issues and the need for a more hybrid approach was highlighted. 
In Miki \emph{et al.} \cite{imageTSP, miki2018applying} was proposed for the first time the use of image processing techniques to solve the TSP.
The choice, even if not obvious in terms of efficiency, has the advantage to get a better understanding of internal network processes.

\section{Materials and Methods}
For the sake of clarity, materials and methods employed in this work have been divided in subsections.
In Section \ref{subsec:2.1}, constructive heuristics, based on fragments growth, are reviewed with some examples. 
The statistical facts and the leading intuitions behind our heuristic are presented in Section \ref{subsec:2.2}.
The general idea of \textit{ML-Constructive} is described in Section \ref{subsec:2.3}, the overall algorithm with its complexity is demonstrated in Section \ref{subsec:2.4}.
The model embodying the ML decision-taker and the training procedure are explained in details in Section \ref{subsec:2.5}.

\subsection{Constructive Heuristics} \label{subsec:2.1}
Constructive heuristics were introduced for the purpose of creating TSP solution from scratch, 
when just the initial formulation described in Section \ref{subsec:1.1} is available.  
They exploit intrinsic features during the solution creation, and are employed when quick solutions with decent quality are needed \cite{ Applegate, multi_fragment, clarke_wright}.

\begin{figure}[!t]
    \centering
    \includegraphics[scale=0.35]{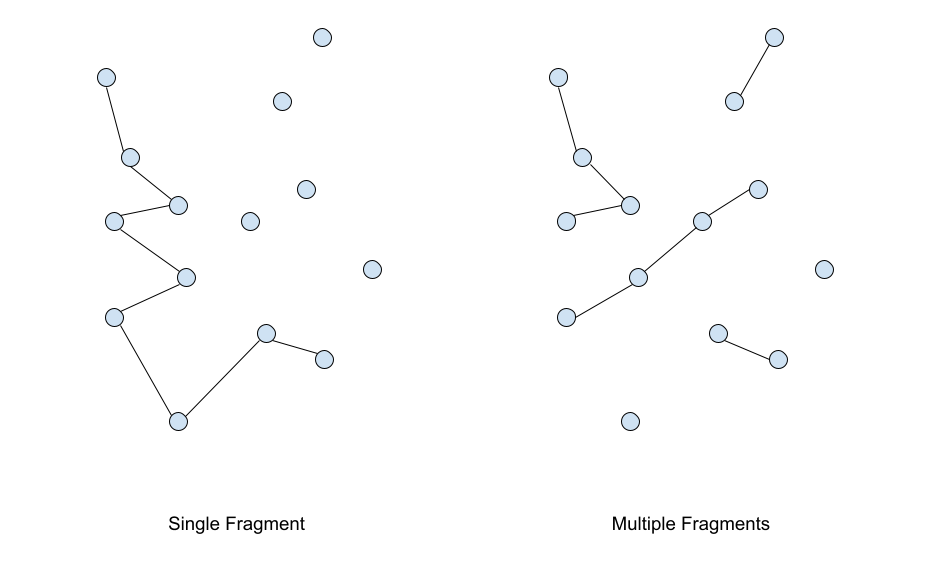}
    \caption{Left Image: single fragment constructor that operates similarly to NN.  Right Image: Constructor that growths multiple fragments similarly to MF and CW. }
    \label{fig:SVvsMF}
\end{figure}

In this paper, we further develop constructive heuristics.
Particularly, those that take their decisions on edges are addressed.
These approaches grow many fragments of tour, in opposition to NN that grows just a single fragment (Figure \ref{fig:SVvsMF}).
They take an extra effort with respect to the latter to preserve the TSP constraint (Equations \ref{b} to \ref{e}).
Since at each addition the procedure must avoid inner-loops and no vertex can be connected with more than two other vertices. 
However, the computational time needed to construct a tour remains limited.    


As introduced in \citet{Wang} and \citet{Jackovich}, we want to emphasize that two main choices are required to design original constructive heuristic driven by edge choices:
the order for the examination of the edges, and the type of constraints that ensure a correct addition.
The total number of edges is $O(n^2)$ for the TSP.
Therefore, the sort of all the edges in the examination order is computed in $O(n^2 \, \log n^2)$. 
The relevance of an edge is related to the probability that we expect that such edge is in the optimal solution.
The higher is the relevance the earlier that edge should be examined.
Different strategies are possible, the most famous ones are the MF and the CW policies.
For MF the relevance of an edge is related to cost values, the smaller is the cost, the higher is the probability of addition.
For CW instead the relevance depends on the saving value, that is designed on purpose to rethink the addition order.
MF sorts the ordering view by the cost values $c_{ij}$, while CW order the examination with a function which computes savings.
A saving is the gain obtained when rather than passing through an hub node $h$ at each step, the salesman uses the straight edge between vertex $i$ and vertex $j$.
The hub vertex $h$ is chosen in such a way that it is at the shortest total distance (TD) from the other vertices.
The formula used to find the hub vertex is in Equation \ref{hub}, while the function that computes the saving values is shown in Equation \ref{eq:Eq2}.

\begin{equation}\label{hub}
    h = \argmin_{i \in V} \; TD[i], \qquad TD[i] = \sum_{j=0}^{n - 1} c_{ij}, \qquad \forall i \in V
\end{equation}

\begin{equation}\label{eq:Eq2}
  s_{ij} = c_{ih} + c_{hj} - c_{ij}, \qquad \forall i,j \in V, \text{ with } i\neq j
\end{equation}

The edge's addition checker algorithm is the second choice in the design of a constructive heuristic.
For both approaches (MF and CW), it checks that the examined edge does not has extremes vertices with already two connections in the partial solution (Equations \ref{b} and \ref{c}), 
and that such edge does not create inner-loops (Equation \ref{d}).
The subroutine that checks if an edge can create an inner-loop is called \emph{tracker},
and it uses a quadratic number of operations for the worst case scenario in our implementation
(Appendix \ref{complexity}).
Note that there exists more efficient data structures and algorithms for the tracker task that runs in $O(n \log n)$ \cite{Jackovich}.

\subsection{Statistical Study} \label{subsec:2.2}

As mentioned earlier, the generalization issues of ML approaches are likely caused by the weak contemplation of the deep learning weaknesses \cite{Bresson, Marcus, DL_limitations}.
In fact, it is known that dealing with imbalanced datasets, outliers and extrapolation tasks can critically effect the overall performances.
An imbalanced dataset occurs when the output classes that the learning model wants to predict present underrepresented and severe class distribution skew \cite{haixiang2017learning, krawczyk2016learning}.
An outlier arises when a data point differs significantly from other observations, outliers can cause problems in their pattern recognition \cite{miller1993tutorial}.
Finally, extrapolation issues arise when the learning system is required to operate beyond the space of known training samples, since the objective is to extend the intrinsic features of the problem to similar but different tasks \cite{bardach2004extrapolation}.

To overcome the aforementioned generalization problems, we suggest supporting the ML model with a designed strategy.
We instruct the model to act as a decision-taker, 
and we decide to place it in a context that allows the ML to act confidently. 
We highlight the significance of designing a good environment that avoids gross errors, i.g. by omitting imbalanced class skews and outlier points.
In fact, choosing wisely a context that does not change too frequently during the algorithm iterations, helps the ML 
to deal with extrapolation skills as well.

As suggested by \citet{Fu}, we address the global problem with subroutines, but rather than treating the ML as an initializing procedure, we value it as decision-taker for the solution construction.
The subroutine task consists of the detection of optimal solution edges from a given CL.
As stated in Section \ref{subsec:1.1}, CL identifies the most promising edges to be part of the optimal solution.
For instance, \citet{Hougardy} proved that 
just around $30 \, n$ of the edges need to be taken into account by an optimal solver for large instances with more than a thousand points. 
Consequently, we want to use an ML decision-taker to recognize the optimal edges from the CL selection. 
Note that for each vertex there is a CL, and for each CL there exists at most two optimal edges. 

\begin{figure}[!t]
    \centering
    \includegraphics[width=0.7\linewidth]{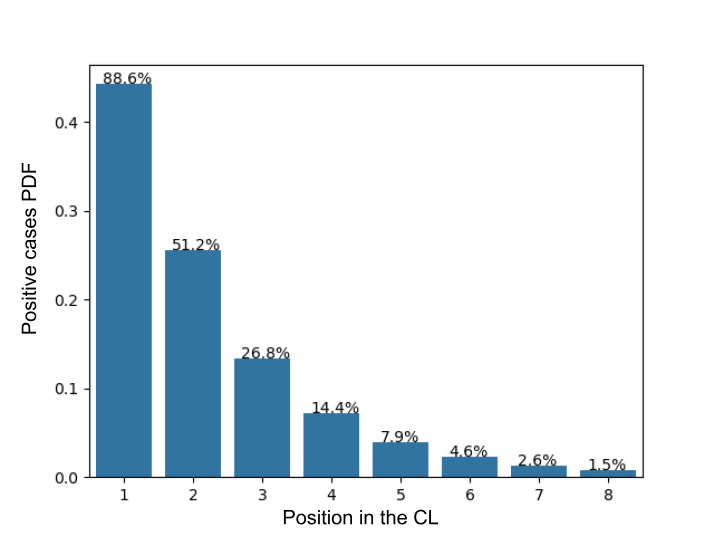}
    \caption{Empirical Probability Density Function (PDF) showing the optimal edge behaviour in relation to the position in the CL. 
    Over each bar is shown the rate of optimal edge occurrence for each considered position.}
    \label{fig:optCL}
\end{figure}

For the purpose of designing good strategies, an explorative study was carried out to check the distribution of the optimal edges within the CLs.
It was observed that after sorting the edges in the CL from the shortest to the longest, the occurring of an optimal edge
is not uniform with respect to the positions in the sorted CL, but follows a logarithmic distribution, as shown in Figure \ref{fig:optCL}.
Clearly, such a pattern reveals a severe class distribution skew for some positions.
In fact, some positions verify the presence of an optimal edge much more infrequently than other positions.

The empirical probability density function (PDF) shown in Figure \ref{fig:optCL} is computed using $1000$ uniform random euclidean TSP instances, with the number of vertices varying uniformly between $100$ and $1000$.
These instances are sampled in the unit side square, and the optimal solutions are computed with the Concorde solver \cite{Applegate}.

Figure \ref{fig:optCL} also shows, in conjunction to the position distribution, the rate of optimal edges found for each position. 
Note that an optimal edge occurrence arises when the optimal tour passes through the edge in the position taken into consideration in the CL.
This study emphasises the relevance of detecting when the CL's shortest edge is optimal, since about $88.6\%$ of the time such an edge is in the optimal tour.
However, it reveals as well that detecting with ML when this edge is not optimal is a hard task due to the over-represented situation. 
Instead, considering the second position (second shortest edge), a more balanced scenario can be observed, since about half of the occurrences are positive and the other half is negative.
A rapid growth of under-represented cases can be observed from the third position onwards, with only few optimal edge occurrences this time.
Note that up to the fifth position the under-representation is not too severe, and imbalanced learning techniques could make their contribution \cite{lemaitre2017imbalanced}.
From the sixth on the optimal occurrences are too rare to be able to recognize useful patterns, even if these cases could be interpreted as the most useful ones in terms of construction.
Considering CLs built in such a way to contain all the vertices in the defined instance, then the number of optimal edges in each CL is two.
And, the sum of optimal occurrences for the first five positions in the CL is about $95\%$ of the total optimal edges available.

\begin{figure}[!t]
\centering
\includegraphics[scale=0.45]{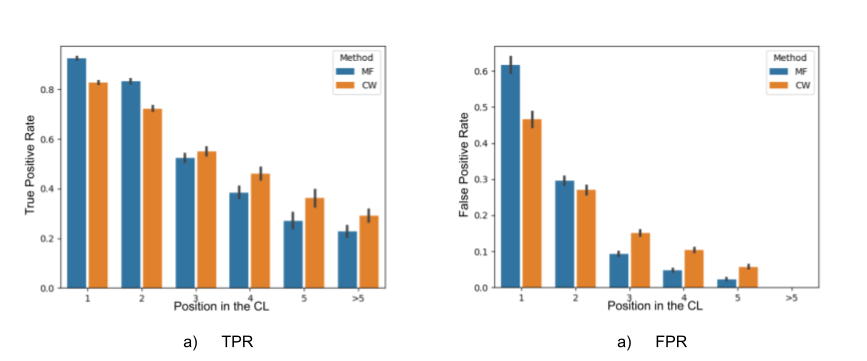}
\caption{ True positive Rate (TPR) and False Positive Rate (FPR) comparison for Multi Fragment (MF) and Clarke Wright (CW). The first five positions in the CL are considered separately, while all the others are shown in the $>5$ bars. }
\label{fig:2}       
\end{figure}

After the selection of the most promising edges to be processed with ML, a relevant choice to be made is the order of examination of these edges.
In fact, as mentioned, edges selected in the earlier stages of the construction exhibit higher probabilities to be inserted regarding the later ones,
since less constraints need to be validated.

To explore the effectiveness of different strategies, we tested the behaviour of classic constructive solvers as MF, and CW (Figure \ref{fig:2}).
Using 54 TSPLIB instances \cite{reinelt1991tsplib} with dimension varying from 100 to 1748 vertices, we investigate the true positive rate (TPR) and the false positive rate (FPR) of these solvers.
Considering each constructive solver as a predictor, each position in the CL as a sample point, and the optimal edge positions as the actual targets to be predicted,
for each CL, there are two position targets and two position predicted.
Considering the distribution shown in Figure \ref{fig:optCL}, the positive (P) and negative (N) cases  occur with a frequency that varies depending on the position.
Hence, studying the predictor performance by position is important since each position has different relevance in the solution construction and optimal frequency.

A true positive occurs when the predicted edge is also an optimal edge, a false positive instead occurs if the predicted edge is not optimal.
Note that avoiding false positive cases is crucial, since they block other optimal edges in the process.
To take care also of this aspect, the positive likelihood rate (PLR) is considered in Table \ref{tab:metrics Classic}.
Let $D_p$ be the dataset of all the edges available in the $p$ positions for each CL.
If the predictor truly find an optimal edge in the observation $i$, the variable $\text{TP}_i$ will be equal to one, otherwise is null.
Similarly, it is for 
the false positive $\text{FP}_i$ variable.

\begin{equation}\label{def_recall_fallout}
    TPR = \frac{ \sum_{i \in D_p} \text{TP}_i}{\sum_{i \in D_p} \text{P}_i},  \qquad FPR = \frac{ \sum_{i \in D_p} \text{FP}_i}{\sum_{i \in D_p} \text{N}_i}, \qquad PLR = \frac{TPR}{FPR}
\end{equation}

In Table \ref{tab:metrics Classic}, MF exhibits an higher TPR for shorter edges as expected, while CW performs better with longer edges.
However, less obvious is MF's higher FPR on the first position.
Bringing attention to the relevance of decreasing the FPR for the most frequent first position.
Note that the CW's PLR is higher with respect to MF's one for the first position, which can be read as the main reason why CW comes up with better TSP solutions than MF.
So one of the main reasons for using ML for TSP is the reduction of the FPR for construction or general heuristics.

\begin{table}[b!]
\centering
\caption{True positive rate (TPR), false positive rate (FPR) and positive likelihood rate (PLR) comparison across several positions and methods.}\label{tab:metrics Classic}
\begin{tabular}{|c|c|cc|c|}\specialrule{.1em}{.0em}{.0em}
position &method &TPR &FPR & PLR \\\specialrule{.1em}{.0em}{.0em}
\multirow{2}{*}{1} &MF &0.9257 &0.6165 &1.50 \\
&CW &0.8279 &0.4656 &1.78 \\\hline
\multirow{2}{*}{2} &MF &0.8321 &0.2957 &2.81 \\
&CW &0.7229 &0.2701 &2.68 \\\hline
\multirow{2}{*}{3} &MF &0.5241 &0.0923 &5.68 \\
&CW &0.5503 &0.1504 &3.66 \\\hline
\multirow{2}{*}{4} &MF &0.3847 &0.0479 &8.03 \\
&CW &0.4599 &0.104 &4.42 \\\hline
\multirow{2}{*}{5} &MF &0.2712 &0.0227 &11.95 \\
&CW &0.3620 &0.0575 &6.30 \\\hline
\multirow{2}{*}{>5} &MF &0.2272 &0.0001 &2272.0 \\
&CW &0.2901 &0.0001 &2901.0 \\
\specialrule{.1em}{.0em}{.0em}
\end{tabular}
\end{table}

\subsection{The general idea} \label{subsec:2.3}
In light of the statistical study presented in Section \ref{subsec:2.2}, we propose a constructive heuristic called \textit{ML-Constructive} (ML-C). 
The heuristic follows the edge addition process (see MF and CW in Section \ref{subsec:2.1}) extended by an auxiliary step that asks the ML model to agree for any attaching edge during a first phase.
The goal is to avoid as much as possible to add bad edges in the solution, while allowing the addition of promising edges which are considered auspicious by the ML model. 
Our focus is not on the development of highly efficient ML architectures, but rather on the successful interaction between machine learning and optimization techniques.
We conceive the machine learning model as a decision-taker, and the optimization heuristics as the texture of the solution building story.
The result is a new hybrid constructive heuristic which succeeds in scaling up to big problems while keeping the improvements achieved through machine learning.

\begin{figure}[b]
    \centering
    \includegraphics[scale=0.25]{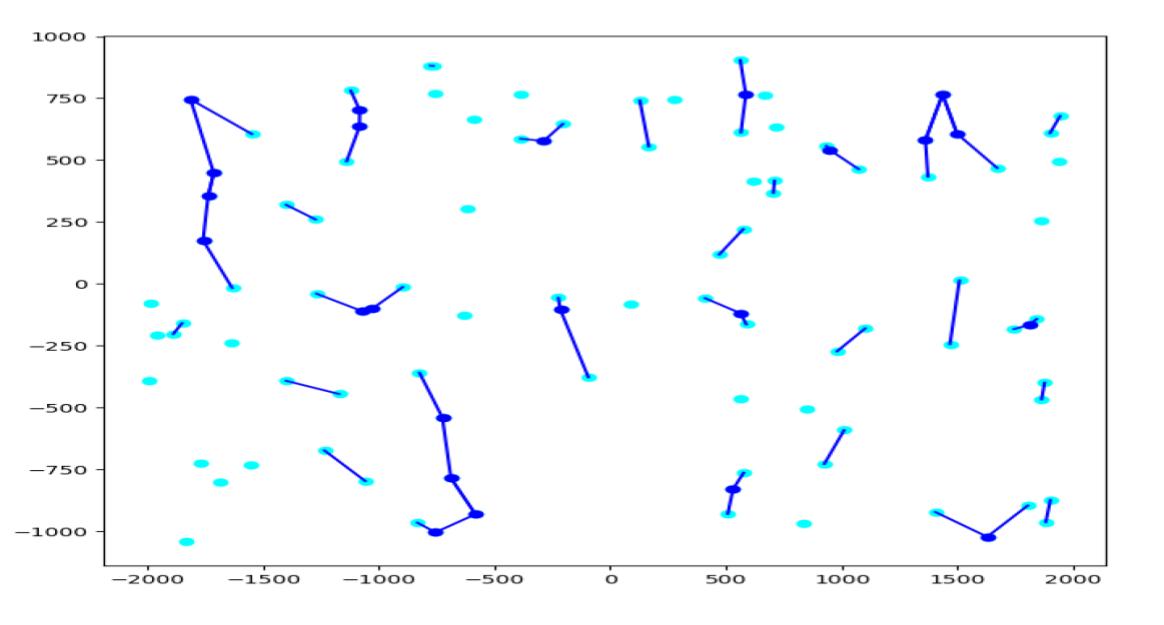}
    \caption{First phase partial solution constructed with the machine learning predictions. Vertices in light blue are free for the second phase of \emph{ML-Constructive}. The instance is the KroA100 from the TSPLIB collection.}
    \label{fig:fragments}
\end{figure}

To avoid the popular ML flaws discussed in previous section, the ML decision-taker is exploited  just on situations where the data do not suggest underrepresented cases.
Since about $95\%$ of the optimal edges are connections with one of the closest five vertices of a candidate list, only such subset of edges is initially considered to test the ML system. 
It is a common practice to avoid employing ML models in the prediction of rare events, instead it is suggested to apply it preferably in cases where a certain degree of pattern recognition can be confidently detected.
Taking this into account, our solution is designed to construct TSP tours in two phases.
The first employs the ML to construct partial solutions where it can predict with confidence (Figure \ref{fig:fragments}).
The second uses the CW heuristic to connect the remaining free vertices and complete the TSP tour.

During the first phase, the most likely edges to be found in the optimal tour are initially collected in the list of promising edges $L_P$.
As mentioned, the edges connecting the first five closest vertices for each CL are assumed as promising for testing.
Note that, as highlighted in Figure \ref{fig:optCL}, from the third shortest edge (position) onwards it is rare to find optimal edges, 
since they are already underrepresented.
Therefore, to appropriately choose the edges to be placed in the promising list, several experiments were carried out and results 
are shown in Table \ref{tab:comparison CL}.
To build the promising list $L_P$ for the experiments, the strategy was to include the edges of the first $m$ vertices of each CL, considering $m$ ranging from 1 to 5.
The ML decision-taker was in charge to predict whether the edges under consideration in $L_P$ (given $m$) were in the optimal solution or not. 
It adopted the same ResNet \cite{resnet} architecture and procedure employed for the \emph{ML-Constructive}, but each ML model was trained on different data in order to be consistent with the $m$ tested case. 
More details on the training data in Section \ref{subsec:2.5}.

Several metrics were compared in the results obtained for the different $m$: 
the True Positive Rate (TPR), the False Positive Rate (FPR), 
the Accuracy (Acc) and the Positive Likelihood Rate (PLR) \cite{colquhoun2017reproducibility}.
Please note that the decision-taker objectives are to keep the FPR small, meanwhile obtain good results in terms of TPR. 
In fact, having small FPR ensures that during the second phase the \emph{ML-Constructive} has an higher probability in detecting optimal edges, while with an high TPR the search space for the second phase is reduced (Appendix \ref{first_probable}). 
The use of these specific metrics could lead to better results.

\begin{table}[!t]
    \centering
    \begin{tabular}{ |c|c c | c c | } 
    \hline
    \hspace{0.2 cm}closest $m$ \hspace{0.2 cm} &\hspace{0.2 cm} TPR \hspace{0.2 cm}&\hspace{0.2 cm} FPR\hspace{0.2 cm}  &\hspace{0.2 cm} Acc \hspace{0.2 cm} & \hspace{0.2 cm} PLR \hspace{0.2 cm}\\
    \hline
    1    & 99.99\% & 99.99\% & 88.6\%& 1.00 \\ 
    2  & 53.91\% & 13.70\% & 63.08\% & 9.19 \\ 
    3  & 30.97\% & 1.64\% & 60.97\% & 18.85 \\
    4  & 31.00\% & 1.50\% & 67.94\% & 20.61 \\
    5  & 38.66\% & 1.30\% & 75.93\% & 29.71 \\
    \hline
    \end{tabular}
    \caption{Comparison on several quality metrics for different choices of the $L_P$ list.  }
    \label{tab:comparison CL}
\end{table}

In terms of accuracy it seems to be the best choice to include only the shortest edge for each CL ($m = 1$), but by checking the TPR and FPR in Table \ref{tab:comparison CL} it becomes obvious that the ML model almost always predicts an insertion for this case.
This behaviour is undesirable as it leads to a high FPR, hence to worse solutions during the second phase.
However, if the difference between TPR and FPR is taken into account, the best arrangement is when the first two shortest edges are put into the list ($m=2$).
Although other arrangements might show to be effective as well, the selection by means of positions in the CLs and the selection of the first two shortest edges in each CL are proven to be efficient by the results.
Recall that too much edges in $L_P$ can confuse the decision-taker, since outlier cases and classes with severe distribution skew can appear.

After the most promising edges have been identified and included in $L_P$, the list is sorted according to an heuristic that seeks to anticipate the processing of good edges.
An edge belonging to the optimal tour and being straightforwardly detected by the ML model is regarded as good.
It is crucial to find an effective sorting heuristic for the promising list, since the order of it effects the learning process and the \emph{ML-Constructive} algorithm as well.
For simplicity, in this work the list is sorted by edge's position in the CL and cost length, but other approaches could be propitious, perhaps using ML.
Note that the earlier examination of the most promising edges increases the probability to find good tours employing the multiple fragment paradigm (Appendix \ref{first_probable}).  

At this point, the edges belonging to the sorted promising list are drawn in images and fed to the ML decision-taker one at a time. 
If the represented edge meets the TSP constraints, the ML system will be challenged to detect if the edge is in the optimal solution.
If the detection is done with a given level of confidence, the heuristic will validate the inclusion of the edge in solution.
Assuming that some local information provides enough details to detect common patterns from previous solutions, the images represent just the small subset of vertices given by the candidate lists of each edge extremes of the edge that is processed.
The partial solution visible in such local context and available up to the insertions made by moving through the promising list is represented as well.

Once all the edges of the promising list have been processed, the second phase of the algorithm is in charge of completing the tour. 
Initially, it detects the remaining free vertices to connect (Figure \ref{fig:fragments}), then it connects such vertices employing a constructive heuristic based on fragments.
Several strategies are possible, but to keep the construction straightforward we choose to conclude the tour with the CW heuristic.
Note that CW usually captures the optimal long edges better than MF, as highlighted in Figure \ref{fig:2} and Table \ref{tab:metrics Classic}.
Therefore CW represents a promising candidate solver to connect the remaining free fragment extremes of the partial solution into the final tour.


\subsection{The ML-Constructive Algorithm}\label{subsec:2.4}

The \emph{ML-Constructive} 
starts as a modified version of MF, then concludes the tour exploiting the CW heuristic.
The ML model behaves consequently as a glue, since it is crucial to determine the partial solution available at the switch between solvers.

The list of promising edges $L_P$ and the confidence level of the ML decision-taker are critical specifications to set in the heuristic before than it runs. 
The reasons behind our 
promising list building choices
were widely discussed in Section \ref{subsec:2.3}.
While, the confidence level is used to handle the exploitation vs exploration trade-off.
It consists in a simple threshold applied to the predictions made by the ML system.
If the predicted probability that validates the insertion is greater than such threshold, then the insertion is applied.
The value of $0.99$ has been verified to provide good results on tested instances.
Since, lower values increase the occurrence of false positive cases, thus leading to the inclusion of edges that are not optimal.
On the other hand, higher values of it decrease the occurrence of true positive cases, hence increasing the challenge of the second phase.

The overall pseudo-code for the heuristic is shown in Algorithm \ref{ML-Constructive}.
Firstly, the candidate lists for each vertex in the instance are computed (line 2). 
We noticed that considering just the closest thirty vertices for each CL was a good option. 
As mentioned, just the first two connections are considered in $L_P$, while the other vertices are used to create the local context in the image.
The CL construction takes a linear number of operation for each vertex, and the overall time complexity for constructing it is $O(n^2)$.
Since finding the nearest vertex of a given vertex it takes linear time, the search for the second nearest takes the same time (after removing the previous from the neighborhood).
So on until the thirtieth nearest vertex is found.
As only the first thirty edges are searched, the operation can be completed in linear time. 
Then, promising edges 
are inserted in $L_P$, meanwhile repeating edges are deleted to avoid unnecessary operations (line 3).
The list is sorted according to the position in the candidate list and the cost values (line 4).
All the edges that are the first nearest will be found first and sorted according to $c_{ij}$, then the second nearest and so on.
Since only the first two edges for each CL can be in the list, the sorting task is completed in $O(n \log n)$.

\begin{algorithm}[!t]
\begin{algorithmic}[1]
\caption{ML-Constructive } \label{ML-Constructive}
\vspace{0.2 cm}
\Require{TSP graph $G(V, E)$}
\Ensure{a feasible tour $X$}
\vspace{0.3 cm}
\Procedure{ML-Constructive}{$G(V, E)$}
\State create CL for each vertex
\State insert the shortest two vertices for each CL into $L_P$ 
\State sort $L_P$ according to the position in the CL and the ascending costs $c_{i,j}$ 

\State $X = \mathbf{\Bar{0}}$ 
\For{ $l$ in $L_P$}
    \State select the extreme vertices $i, j$ of $l$
    \If{ vertex $i$ \textbf{and} vertex $j$ have exactly one connection each in $X$}:
    \If{$l$ do not creates a inner-loop}:
    \If{the ML agrees the addition of $l$}: $\, x_{i, j}=1$\;
    \EndIf
    \EndIf
    \Else
    \If{ vertex $i$ \textbf{and} vertex $j$ have less than two connections each in $X$}:
    \If{the ML agrees the addition of $l$}: $\, x_{i, j}=1$\;
    \EndIf
    \EndIf
    \EndIf
\EndFor
\State find the hub vertex $h$
\State select all the edges that connects free vertices and insert them into $L_D$
\State compute the saving values with respect to $h$ for each edge in $L_D$ 
\State sort $L_D$ according to the descending savings $s_{i, j}$  
\State $t = 0$\;
\While{the solution $X$ is not complete}
    \State $l = L_D[t]$,  \hspace{0.2 cm} $t = t + 1$ \;  
    \State select the extreme vertices $i, j$ of $l$
    \If{ vertex $i$ \textbf{and} vertex $j$ have exactly one connection each in $X$}:
    \If{$l$ do not creates a inner-loop}:  $\, x_{i, j}=1$\;
    \EndIf
    \Else
    \If{ vertex $i$ \textbf{and} vertex $j$ have less than two connections each in $X$}: 
    \State $x_{i, j}=1$\;
    \EndIf
    \EndIf
\EndWhile
\EndProcedure
\end{algorithmic}
\end{algorithm}

The first phase of \textit{ML-Constructive} takes part.
An empty solution 
$X= \Bar{0} $ 
is initialized (line 5), 
 and following the order in $L_P$ a variable $l$ is updated with the edge considered for the addition (line 6).
At first, $l$ is checked to ensure that the edge complies the TSP constraints (Equations \ref{b} to \ref{d}).
Then the ML decision-taker is queried to confirm the addition of the edge $l$.
If the predicted probability is higher than the confidence level, the edge is added to the partial solution (lines 10 and 13).
To evaluate the number of operations that this phase consumes, we must split the task according to the various sub-routines
which are acted at each new addition in solution.
The \enquote{if} statements (lines $8$, $12$, $22$ and $25$) check that the constraints \ref{b} and \ref{c} are complied.
They verify that both extremes of the attaching edge $l$ exhibit at most one connection in the current solution. 
The operation is computed with the help of hash maps, and it takes constant time for each considered edge $l$. 
The tracker verification (lines $9$ and $23$) ensures that $l$ will not create an inner-loop (Equation \ref{d}). 
This sub-routine is applied only after that is checked that both extremes of $l$ have exactly one connection each in the current partial solution.  
It takes overall $O(n^2)$ operations up to the final tour (proof in Appendix \ref{complexity}).

Once all the constraints of the TSP have been met,
the edge $l$ is processed by the ML decision-taker (lines $10$ and $13$).
Even if time consuming, such sub-routine is completed
in constant time for each $l$. 
Initially the image depicting the $l$ edge and its local information is created, then it is given as input to the neural network.
To create the image, the vertices of the candidate lists and the existing connections in the current partial solution must be retrieved.
Hash maps are used for both tasks, and since the image can include up to sixty vertices, this operation takes a constant amount of time for each $l$ edge in $L_P$.
The size of the neural network does not vary with the number of vertices in the problem as well, but remains constant for each edge in $L_P$.

To complete the tour, the second phase starts by identifying the hub vertex (line 14). 
It consider all the vertices in the problem (free and not), following the rule explained in Equation \ref{hub}. 
Free vertices are selected from the partial solution, and edges connecting such vertices are inserted in the difficult edges list $L_D$ (line 15). 
The saving for each edge in $L_D$ is computed (line 16), 
and the list is sorted according to these values (line 17) in $O(n^2 \log n^2)$. 
At this point (lines 19 to 26), the solution is completed employing the classical multiple fragment steps, which are known to be $O(n^2)$ \cite{multi_fragment}.

Therefore, the complexity of the worst case scenario for the \emph{ML-Constructive} is:

\begin{equation}
 O(n + n \log n  + n^2 + n^2 \log n^2) = O(n^2 \log n^2)   
\end{equation}

Note that to complete the tour we proposed the use of CW, 
but rather hybrid approaches that also use some sort of exhaustive search 
could be very promising as well, although more time consuming.


\subsection{The ML decision-taker}\label{subsec:2.5}
The ML decision-taker validates 
the insertions made by the \emph{ML-Constructive} during the first phase.
Its scope is to exploit the ML pattern recognition to increase the occurrences of finding good edges, while reducing them
for the bad edges.


Two data-sets were specifically created to fit the ML decision-taker, the first was used to train the ML system while the second to evaluate and choose the best model.
The training data-set is composed by $38400$ instances uniformly sampled in the unit-sided square.
The total number of vertices for instance $n$ ranges uniformly from 100 to 300.
On the other hand, the evaluation data-set is composed by 1000 instances uniformly sampled from the unit-sided square.
The total number of vertices varies in this case from 500 to 1000.
The data-set has been used 
to create the results in Table \ref{tab:comparison CL}.
The optimal solutions were found (in both cases) using the Concorde solver \cite{Applegate}.
The creation of the training instances and their optimal tours took about 12 hours on a single CPU thread;
while a total time of 24 CPU hours was needed for the evaluation data-set creation (since it includes instances of greater size).
Note that, in comparison with other approaches that use 
Reinforcement Learning
good results were achieved here even though we used far fewer training instances \cite{Mele1}.

To get the ML input ready, the promising list $L_P$ was created for each instance in the data-sets.
In case $m=2$ (best scenario), the two shortest edges of each candidate list were inserted into the list.
To avoid repetitions, edges that occur several times in the list were inserted just once at the shortest available position.
For example, if edge $e_{ij}$ is the first shortest edge in CL$[i]$ and the second shortest edge in CL$[j]$, it will only occur in $L_P$ once such as first position for vertex $i$.
After that all promising edges had been inserted into $L_P$, the list was sorted.
Note that the list can contain at most $m\times n$ items in it.

\begin{figure}[t!]
    \centering
    \includegraphics[scale=0.25]{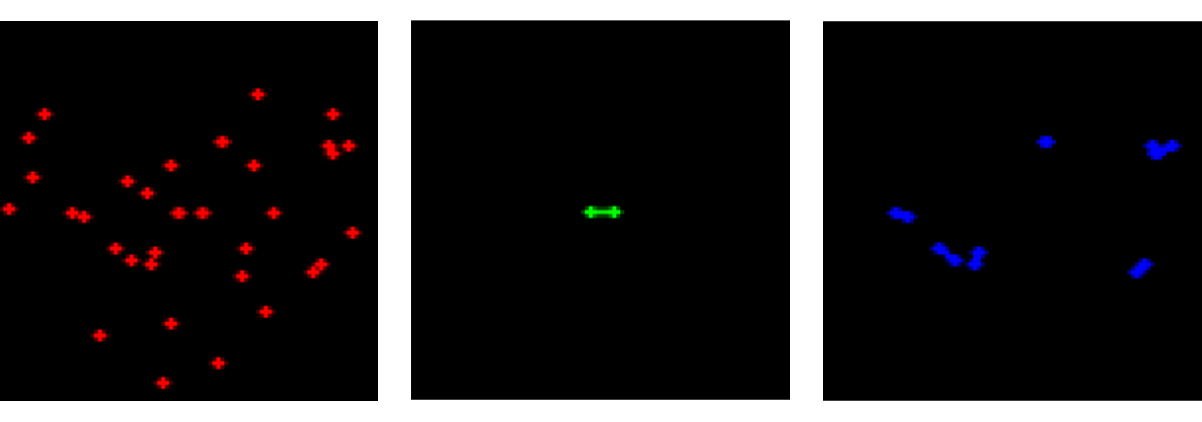}
    \caption{Example of input image. The vertices in the local view are in the red channel, the $l$ edge is drawn in green channel, while the edges in the partial solution are in the blue channel.}
    \label{fig:input images}
\end{figure}

An image with dimension of $96\times 96 \times 3$ pixels was created for each edge belonging to $L_P$. 
Three channels (red, green and blue) were set up to provide the information used to feed into the neural network.
Each channel depicts some information inside a square with sides of 96 pixels each (Figure \ref{fig:input images}).
The first channel of the image (red) shows each vertex in the local view,
the second one (green) shows the edge $l$ considered for the insertion with its extremes,
and the third one (blue) shows all the existing fragments currently in the partial solution and visible from the local view drawn in the first channel.

As mentioned, the local view was formed by merging the vertices belonging to the candidate lists of each extremes of the inserting edge.
These vertices were collected and their positions normalized to fill the image.
The normalization required having the middle of the inserting edge $l$
such as image center.
Whereas, all the vertices visible in the local view were interior to a virtual sphere inscribed into the squared image.
Such that the maximum distance between the image center and the vertices in the local view was less than the radius of such sphere.
The scope of the normalization was to keep consistency among the images created for the various instances. 

The third channel was concerned in giving a temporal indication to aid the ML system in its decision.
In fact, representing those edges which had been inserted during the previous stages of the \emph{ML-Constructive}, 
this information gave a helpful hint in the interpretation of which edges the final solution needs most.
Two different policies were employed in the construction of it: the optimal policy (offline) and the ML adaptive policy (online).
The first used the optimal tour and the $L_P$ order to create this channel (just on training), while the second one used the ML previous validations (train and test).



\newpage

\begin{wrapfigure}{l}{4.5cm}
\includegraphics[scale=0.65]{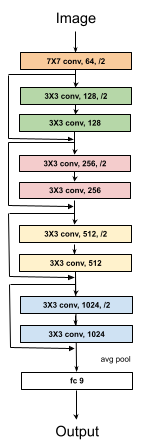}
\caption{ResNet10}
\label{fig:ResNet}
\end{wrapfigure}

A simple ResNet \cite{resnet} with 10 layers was adopted to agree on the inclusion of the edges into the solution.
The choice of the model is motivated by the easiness that image processing ML models show on the understanding of the learning process.
The architecture is shown in Figure \ref{fig:ResNet}.
There are four residual connections, containing two convolutional 
layers each.
The first layer in each residual connection is characterized by a stride equal to two (/2).
As usual for the ResNet the kernels are set to 3x3, and the number of features increase by multiplying by 2 at each residual connection, in order to balance the downscale of the images.
The output is precede by a fully connected layer with 9 neurons (fc) and by an average pool operation (avg pool), that shrinks the information in a single vector with 1024 features.
For additional details we refer to \citet{resnet}.
The model is very compact, 
with the scope of avoiding computational burden and other complexities.

The output of the network is represented by two neurons.
One neuron is predicting the probability that the considered edge $l$ is in the optimal solution, while the other is predicting if the edge is not optimal.
The sum of both probabilities is equal to one.
The choice of using two neurons as output instead of just one is due to the exploitation of the Cross Entropy loss function, which is 
recommended to train classification problems.
In fact, this loss penalizes especially those predictions that are confident and wrong. 
The network will know if the inserting edge $l$ is optimal or not during train, while the ResNet should predict the probability that the edge is optimal during test.

To train the network two loss functions were jointly utilized: the Cross Entropy loss (Equation \ref{CE}) \cite{de2005tutorial} and a reinforcement loss (Equation \ref{reinforce}) which was developed specifically for the task at hand.
Initially, the first loss is employed alone up to convergence (about 1000 back-prop iterations), then the second loss is also engaged in the training.
At each iteration of back-propagation the first loss updates the network firstly, then (after 1000 iterations) the second loss updates the network as well.
The gradient of the second loss function is approximated by the REINFORCE algorithm \cite{williams1992simple} with a simple moving average with 100 periods used as baseline.

\begin{eqnarray}
    \textrm{loss}_1 &=& -  \mathop{\mathbb{E}}_{p(x_l)} \big[ \log q_{\theta}(x_l) \big]\label{CE}\\ 
    \textrm{loss}_2 &=& - \mathop{\mathbb{E}}_{q_{\theta}(x_l)} \big[ T(x_l) - F(x_l) \big] \label{reinforce}
\end{eqnarray}
    
\noindent In Equations \ref{CE} and \ref{reinforce}, $x_l$ is the image of the inserting edge $l$, the function identifying whether $l$ is optimal is accounted as $p$, while $q_{\theta}$ is the ResNet approximation to it. 
Moreover, the $T$ function returns one if the prediction made by $q_{\theta}$ is true (TP or TN), and zero otherwise.
While the $F$ function returns one if the prediction is false.
Note that the second loss exhibit an expected value with respect to $q_{\theta}$ measure, since the third channel is updated using the the ML adaptive policy (online), while the first loss uses the optimal policy (offline).
The introduction of a second loss had the purpose of increasing the occurrences of true positive while decreasing the false positive cases. Note that it employs the same policy that will occur during the \emph{ML-Constructive} test run.




\section{Experiments \& Results} \label{sec:4}

To test the efficiency of the proposed heuristic, experiments were carried out on 54 standard instances.
Such instances were taken from the TSPLIB collection \cite{reinelt1991tsplib}, and
their vertex set cardinality vary from 100 to 1748 vertices.
Non-euclidean instances, such as the ones involving geographical distances (GEO) or special distance functions (ATT),  were addressed as well.
We recall that the ResNet model was trained on small (100 to 300 vertices) uniform random euclidean instances, evaluated on medium-large (500 to 1000 vertices) uniform random euclidean instances, and tested on TSPLIB instances.
We emphasize that TSPLIB instances are generally not uniformly distributed.

All the experiments\footnote{All the code for the experiments replication and for the data-sets creation can be found in the github repository: \url{https://github.com/UmbertoJr/ML-Constructive}
} were handled employing python 3.8.5 \cite{van1995python} for the algorithmic component, and pytorch 1.7.1 \cite{paszke2017automatic} to manage the neural networks.
The following hardware was utilized:
\begin{itemize}
    \item a single {GPU} \textit{NVIDIA GeForce GTX 1050 Max-Q};
    \item a single {CPU} \textit{Intel(R) Core(TM) i7-8750H @ 2.20GHz}.
\end{itemize}
During training all hardware was exploited, while just the CPU was used to test.

The experiments presented in Table \ref{tab:1} compare \emph{ML-Constructive} (ML-C) results to other famous strategies based on fragments,
such as the Multi-Fragment (MF) and the Clarke-Wright (CW).
The first is equivalent to including all the existing edges in the list of promising edges $L_P$, and then substituting the ML decision taker with a rule that always inserts the considered edge.
While, the second strategy is equivalent to keeping the list $L_P$ empty; 
this means that the first phase of \emph{ML-Constructive} does not creates any fragment, while the construction is made completely in the second phase.

\begin{table}[!b]\centering
\caption{Percentage error comparison of various decision-takers policies for 54 testing TSPLIB instances}\label{tab:1}
\scriptsize
\begin{tabular}{lccccccccccccc}\toprule
\textbf{Instances} &\textbf{MF} &\textbf{CW} &\textbf{F} &\textbf{S} &\textbf{Y} &\textbf{AE} &\textbf{BE} &\textbf{ML-C} &  \hspace{0.2cm} \textbf{ML-SC} &\textbf{gap} \\
\noalign{\smallskip}
\hline
\noalign{\smallskip}
kroA100 &14.120 &\textbf{6.043} &9.618 &22.437 &8.636 &11.986 &7.429 &6.480 &  \hspace{0.2cm} \textbf{3.792} &2.251 \\
kroC100 &12.270 &11.480 &8.362 &27.558 &\textbf{5.263} &13.391 &6.950 &10.343 &  \hspace{0.2cm} \textbf{4.776} &0.487 \\
rd100 &16.928 &8.736 &11.580 &24.121 &11.214 &14.212 &8.938 &\textbf{8.559} &  \hspace{0.2cm} 6.738 &1.821 \\
eil101 &27.504 &5.087 &8.426 &22.099 &18.760 &14.348 &8.426 &\textbf{4.293} &  \hspace{0.2cm} \textbf{0.000} &4.293 \\
lin105 &16.065 &8.638 &7.177 &37.332 &21.406 &12.580 &\textbf{7.080} &8.485 &  \hspace{0.2cm} 10.780 &-3.700 \\
pr107 &\textbf{5.799} &10.166 &9.245 &10.640 &8.936 &9.454 &6.717 &11.153 &  \hspace{0.2cm} \textbf{0.445} &5.354 \\
pr124 &10.110 &\textbf{2.502} &4.911 &24.344 &7.942 &9.079 &4.191 &6.991 &  \hspace{0.2cm} \textbf{2.997} &-0.495 \\
bier127 &14.186 &5.659 &4.753 &25.162 &12.370 &10.091 &\textbf{4.571} &6.604 &  \hspace{0.2cm} \textbf{0.000} &4.571 \\
ch130 &28.462 &7.480 &7.414 &22.733 &8.003 &12.305 &8.429 &\textbf{4.975} &  \hspace{0.2cm} \textbf{4.206} &0.769 \\
pr136 &23.160 &\textbf{7.186} &11.709 &15.693 &16.046 &14.878 &11.701 &11.151 &  \hspace{0.2cm} \textbf{3.713} &3.473 \\
gr137 &27.234 &8.243 &9.742 &30.207 &15.548 &14.170 &10.124 &\textbf{7.329} &  \hspace{0.2cm} \textbf{3.188} &4.141 \\
pr144 &12.483 &6.444 &6.628 &12.016 &4.161 &7.625 &\textbf{3.796} &4.474 &  \hspace{0.2cm} \textbf{3.962} &-0.166 \\
kroA150 &20.238 &8.468 &10.507 &26.934 &14.134 &12.799 &9.467 &\textbf{6.877} &  \hspace{0.2cm} \textbf{1.139} &5.738 \\
pr152 &15.196 &9.455 &7.204 &17.060 &6.117 &8.239 &\textbf{5.647} &6.919 &  \hspace{0.2cm} \textbf{3.793} &1.854 \\
u159 &17.952 &8.408 &9.009 &19.881 &10.542 &11.788 &\textbf{5.589} &7.952 &  \hspace{0.2cm} \textbf{6.024} &-0.435 \\
rat195 &13.043 &\textbf{5.854} &7.576 &13.431 &13.345 &10.286 &\textbf{5.854} &7.533 &  \hspace{0.2cm} \textbf{0.000} &5.854 \\
d198 &20.507 &\textbf{5.444} &6.711 &17.535 &7.744 &9.262 &6.267 &6.255 &  \hspace{0.2cm} \textbf{4.011} &1.433 \\
kroA200 &17.819 &8.622 &11.097 &25.541 &11.298 &13.008 &10.328 &\textbf{6.681} &  \hspace{0.2cm} \textbf{2.094} &4.587 \\
gr202 &15.935 &5.683 &7.367 &19.916 &7.716 &9.673 &6.331 &\textbf{4.436} &  \hspace{0.2cm} \textbf{1.825} &2.611 \\
ts225 &12.842 &6.804 &9.493 &6.975 &14.929 &11.309 &8.075 &\textbf{6.520} &  \hspace{0.2cm} 11.330 &-4.810 \\
tsp225 &26.237 &10.438 &9.790 &24.165 &9.609 &13.906 &\textbf{9.169} &11.292 &  \hspace{0.2cm} \textbf{4.403} &4.766 \\
pr226 &21.052 &9.948 &11.918 &16.784 &9.031 &12.347 &8.778 &\textbf{8.599} &  \hspace{0.2cm} \textbf{5.370} &3.229 \\
gr229 &19.624 &8.849 &7.593 &22.242 &10.932 &10.807 &\textbf{6.573} &7.495 &  \hspace{0.2cm} \textbf{2.807} &3.766 \\
gil262 &12.279 &9.714 &6.602 &24.769 &10.892 &10.976 &8.368 &\textbf{6.224} &  \hspace{0.2cm} 8.999 &-2.775 \\
pr264 &14.987 &8.839 &\textbf{4.919} &16.239 &9.816 &10.091 &6.195 &6.036 &  \hspace{0.2cm} \textbf{3.739} &1.180 \\
a280 &20.822 &13.998 &13.959 &15.394 &12.447 &15.944 &12.330 &\textbf{11.439} &  \hspace{0.2cm} \textbf{0.388} &11.051 \\
pr299 &21.639 &\textbf{8.103} &10.467 &22.371 &15.244 &14.385 &10.965 &8.636 &  \hspace{0.2cm} \textbf{0.905} &7.198 \\
lin318 &18.356 &7.849 &\textbf{6.377} &30.848 &14.114 &12.652 &9.398 &6.679 &  \hspace{0.2cm} \textbf{5.501} &0.876 \\
rd400 &15.032 &9.548 &8.278 &21.923 &10.844 &11.840 &\textbf{7.506} &8.108 &  \hspace{0.2cm} \textbf{3.423} &4.083 \\
fl417 &12.469 &12.335 &10.606 &28.488 &8.962 &12.149 &8.473 &\textbf{8.102} &  \hspace{0.2cm} \textbf{7.394} &0.708 \\
gr431 &19.672 &11.953 &\textbf{6.942} &21.114 &11.550 &12.270 &7.867 &11.554 &  \hspace{0.2cm} \textbf{4.196} &2.746 \\
pr439 &15.983 &14.609 &9.024 &23.183 &13.395 &12.157 &9.439 &\textbf{7.661} &  \hspace{0.2cm} \textbf{7.104} &0.557 \\
pcb442 &21.423 &\textbf{9.935} &12.460 &16.403 &13.908 &13.233 &11.596 &10.172 &  \hspace{0.2cm} \textbf{2.787} &7.148 \\
d493 &16.550 &8.652 &9.618 &17.898 &9.872 &9.749 &8.443 &\textbf{6.818} &  \hspace{0.2cm} \textbf{4.352} &2.466 \\
\textbf{att532} &22.223 &11.013 &\textbf{7.596} &25.022 &11.150 &11.629 &9.455 &8.441 &  \hspace{0.2cm} \textbf{3.590} &4.006 \\
u574 &22.276 &10.779 &\textbf{9.023} &24.349 &9.638 &13.113 &9.776 &9.790 &  \hspace{0.2cm} \textbf{4.495} &4.528 \\
rat575 &18.529 &8.460 &10.350 &19.592 &8.962 &11.734 &9.671 &\textbf{6.201} &  \hspace{0.2cm} \textbf{2.761} &3.440 \\
d657 &13.997 &7.949 &\textbf{7.583} &23.008 &8.878 &12.009 &10.043 &7.587 &  \hspace{0.2cm} \textbf{3.829} &3.754 \\
gr666 &13.473 &13.241 &\textbf{9.314} &24.339 &12.670 &13.736 &11.534 &9.635 &  \hspace{0.2cm} \textbf{6.741} &2.573 \\
u724 &17.836 &9.881 &7.590 &23.028 &9.952 &11.276 &9.308 &\textbf{6.543} &  \hspace{0.2cm} \textbf{3.054} &3.489 \\
rat783 &22.062 &9.642 &7.047 &21.549 &11.617 &10.995 &8.877 &\textbf{5.934} &  \hspace{0.2cm} \textbf{4.956} &0.978 \\
pr1002 &18.857 &10.763 &9.751 &20.484 &13.648 &13.238 &11.600 &\textbf{8.529} &  \hspace{0.2cm} \textbf{5.364} &3.165 \\
u1060 &17.322 &10.732 &9.620 &21.754 &10.537 &13.128 &11.580 &\textbf{8.954} &  \hspace{0.2cm} \textbf{6.537} &2.417 \\
vm1084 &23.083 &10.298 &9.615 &28.746 &12.485 &13.390 &11.253 &\textbf{9.123} &  \hspace{0.2cm} \textbf{6.951} &2.172 \\
pcb1173 &17.792 &10.917 &\textbf{9.567} &21.033 &14.631 &13.380 &11.821 &9.986 &  \hspace{0.2cm} \textbf{5.790} &3.777 \\
d1291 &21.917 &10.155 &\textbf{5.711} &14.783 &9.653 &9.643 &7.573 &8.535 &  \hspace{0.2cm} \textbf{3.081} &2.630 \\
rl1304 &12.142 &10.610 &\textbf{7.512} &25.060 &11.100 &11.821 &9.580 &9.506 &  \hspace{0.2cm} \textbf{5.228} &2.284 \\
rl1323 &14.876 &11.804 &7.467 &25.401 &8.385 &11.310 &9.789 &\textbf{6.538} &  \hspace{0.2cm} \textbf{3.695} &2.843 \\
nrw1379 &22.314 &9.914 &8.842 &19.794 &11.351 &11.443 &9.856 &\textbf{7.996} &  \hspace{0.2cm} \textbf{3.655} &4.341 \\
fl1400 &20.520 &11.432 &10.975 &27.267 &13.718 &14.541 &\textbf{9.997} &11.273 &  \hspace{0.2cm} \textbf{8.471} &1.526 \\
u1432 &23.329 &10.407 &12.676 &14.967 &14.928 &14.669 &12.491 &\textbf{8.440} &  \hspace{0.2cm} \textbf{5.725} &2.715 \\
fl1577 &16.976 &12.167 &12.634 &14.082 &12.351 &12.502 &\textbf{9.084} &10.463 &  \hspace{0.2cm} \textbf{4.373} &4.711 \\
d1655 &15.282 &11.187 &9.503 &18.427 &10.368 &12.130 &9.478 &\textbf{8.269} &  \hspace{0.2cm} \textbf{5.637} &2.632 \\
vm1748 &14.134 &11.912 &9.992 &24.520 &11.868 &13.757 &12.225 &\textbf{9.302} &  \hspace{0.2cm} \textbf{6.094} &3.208 \\
\noalign{\smallskip}
\hline
\noalign{\smallskip}
\textbf{average} &\textbf{17.906} &\textbf{9.341} &\textbf{8.879} &\textbf{21.493} &\textbf{11.345} &\textbf{12.082} &\textbf{8.815} &\textbf{8.035} &  \hspace{0.2cm} \textbf{4.374} &\textbf{2.737} \\
\textbf{std} &\textbf{4.659} &\textbf{2.368} &\textbf{2.077} &\textbf{5.493} &\textbf{3.142} &\textbf{1.789} &\textbf{2.155} &\textbf{1.875} &  \hspace{0.2cm} \textbf{2.473} &\textbf{2.600} \\
\textbf{best} &1/54 &7/54 &13/54 &0/54 &4/54 &0/54 &11/54 &29/54 &  \hspace{0.2cm} 50/54 & \\
\bottomrule
\end{tabular}
\end{table}

\begin{figure}[!t]
    \centering
    \includegraphics[scale = 0.4]{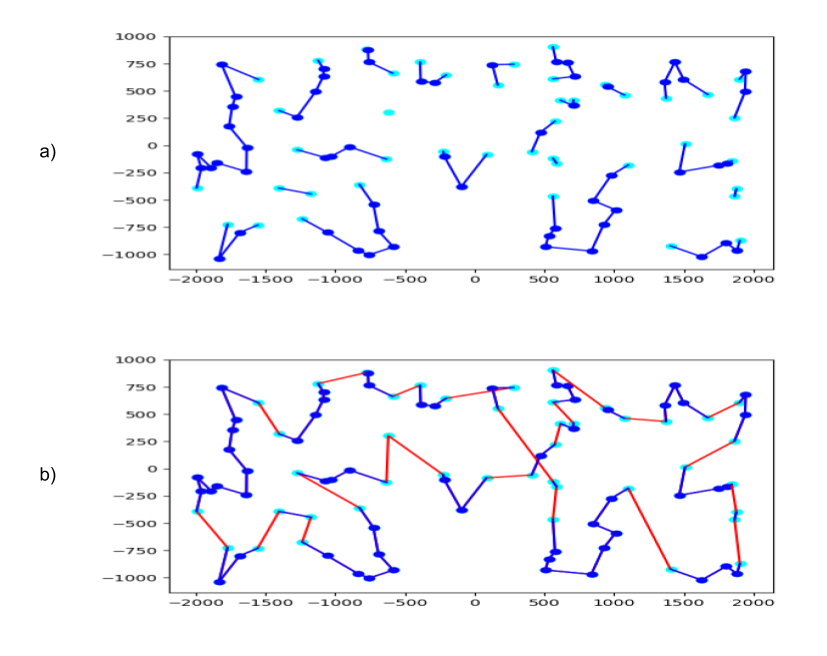}
    \caption{\textbf{a}) The partial solution available at the end of the first phase for ML-SC. In light blue the remaining free vertices, and in dark blue the inserted edges.  
    \textbf{b}) The complete tour found at the end of the ML-SC run. In red the edges added during the second phase. The considered instance is the KroA100 from the TSPLIB collection.}
    \label{fig:opt fragments}
\end{figure}

To explore, evaluate and interpret the behaviour of our two-phase algorithm, other strategies were investigated as well.
In fact, the ML decision-taker can act in very different ways, and a comparison with expert-made heuristic rules can be significant. 
Deterministic and stochastic heuristic rules were created to explore the optimality gap variation. 
The aim was to prove that the learnt ML model would produce a higher gain with respect to the heuristic rules, as corroborated by Table \ref{tab:1}.
The heuristic rules were substituting the ML decision-taker component within the \emph{ML-Constructive} (lines 10 and 13 of Algorithm \ref{ML-Constructive}).
No changes on the selecting and sorting strategies were applied to create the lists $L_P$ and $L_D$.  
The First (F) rule decides to deterministically add the $l$ edge if one of its extremes is the first closest vertices in the CL of the other extreme.
The Second (S) rule is similar, but it adds $l$ only if one extreme is the second shortest in the CL of the other.
The policy that always validates (Y) the insertion of the edges in $L_P$ 
was examined as well.
It represents with CW the extreme cases where the ML decision-taker always validates or not, respectively, the edges in $L_P$.
A stochastic strategy called empirical (E) was tested as well, which adds the edges in $L_P$ according to the distribution seen in Figure \ref{fig:optCL}.
Therefore, it inserts the edge if one of the extremes is the first with probability $0.886$ or the second with probability $0.512$. 
Twenty runs of the empirical strategy were made, and in (AE) we show the average results, while in (BE) we show the best from all runs.
Finally, to check the behaviour of the \textit{ML-Constructive} in case the ML system validates with $100\%$ accuracy (the ML decision-taker is a perfect oracle), the Super Confident (ML-SC) policy was examined.
This policy 
always answers correctly for all the edges in the promising list $L_P$, and is achieved by exploiting the known optimal tours.  
To capture the potentiality of a super accurate network for the first phase, 
the partial solution created by the ML-SC policy in the first phase, and the solution constructed in the second phase are shown in Figure \ref{fig:opt fragments}.  
Note that some crossing edges are created in the second phase.
In fact, despite the solution created being very close to the optimal, the second phase sometimes adds bad edges to the solution.
This last artificial policy has been added to demonstrate how much leverage we can still gain from the machine learning point of view.




The results obtained by the optimal policy (ML-SC) lead us to two interesting aspects.
The first one, as mentioned before, shows the possible leverage from the ML perspective (first phase).
While the second one gives us an idea of how much improvement is possible from the heuristic point of view (second phase).
To emphasize these aspects the gap column in Table \ref{tab:1} clearly highlights the difference, in terms of percentage error, between the ML-SC solution and the best solution found by the other heuristics (in bold). 
The average, the standard deviation (std) and the number of times when the heuristic is best are shown as well for each strategy.

Among the many policies shown in Table \ref{tab:1}, the First (F) and the ML based (ML-C) policies exhibit comparable average gaps.
To prove that the enhancement introduced by the ML system is on average statistically significant, a statistical test was conducted.
A T-test on the percentage errors obtained for the 54 instances in Table \ref{tab:1} shows that the p-value against the hypothesis both policies are similar in terms of average optimal gap is equal to 0.03.
The result proves that the enhancement 
is relevant, 
and that these systems have a promising role in improving the quality of TSP solvers.

\begin{table}[!b]\centering
\caption{CPU time comparison related to different greedy policies for 54 TSPLIB instances}\label{tab:time}
\scriptsize
\begin{tabular}{lrrrrrrrrrr}\toprule
\textbf{Instances} &\textbf{MF} &\textbf{CW} &\textbf{F} &\textbf{S} &\textbf{Y} &\textbf{AE} &\textbf{BE} &\textbf{ML-C} &\textbf{ML-SC} \\
\noalign{\smallskip}
\hline
\noalign{\smallskip}
kroA100 &\textbf{0.004} &0.006 &0.009 &0.008 &0.007 &0.014 &0.276 &1.423 &0.011 \\
kroC100 &\textbf{0.004} &0.006 &0.011 &0.010 &0.008 &0.015 &0.309 &1.706 &0.010 \\
rd100 &\textbf{0.004} &0.008 &0.011 &0.010 &0.008 &0.016 &0.323 &1.608 &0.011 \\
eil101 &\textbf{0.006} &0.007 &0.012 &0.011 &0.011 &0.019 &0.371 &2.178 &0.013 \\
lin105 &\textbf{0.005} &0.007 &0.013 &0.012 &0.011 &0.017 &0.335 &1.445 &0.012 \\
pr107 &\textbf{0.005} &0.009 &0.013 &0.011 &0.011 &0.015 &0.302 &1.298 &0.012 \\
pr124 &\textbf{0.005} &0.009 &0.017 &0.017 &0.016 &0.020 &0.395 &1.565 &0.015 \\
bier127 &\textbf{0.010} &0.012 &0.100 &0.097 &0.047 &0.055 &1.106 &2.554 &0.059 \\
ch130 &\textbf{0.011} &0.011 &0.017 &0.016 &0.015 &0.024 &0.473 &2.507 &0.019 \\
pr136 &\textbf{0.009} &0.011 &0.015 &0.014 &0.014 &0.022 &0.438 &2.308 &0.017 \\
gr137 &\textbf{0.006} &0.012 &0.023 &0.021 &0.020 &0.027 &0.547 &2.557 &0.024 \\
pr144 &\textbf{0.006} &0.015 &0.023 &0.022 &0.018 &0.024 &0.487 &0.640 &0.021 \\
kroA150 &\textbf{0.011} &0.012 &0.021 &0.017 &0.017 &0.025 &0.492 &2.444 &0.019 \\
pr152 &\textbf{0.010} &0.015 &0.023 &0.021 &0.019 &0.026 &0.528 &1.047 &0.022 \\
u159 &\textbf{0.007} &0.015 &0.026 &0.025 &0.024 &0.029 &0.580 &2.926 &0.026 \\
rat195 &\textbf{0.009} &0.024 &0.027 &0.027 &0.026 &0.038 &0.765 &3.553 &0.025 \\
d198 &\textbf{0.022} &0.027 &0.111 &0.109 &0.106 &0.115 &2.305 &3.451 &0.106 \\
kroA200 &\textbf{0.011} &0.028 &0.036 &0.032 &0.024 &0.044 &0.879 &3.975 &0.027 \\
gr202 &\textbf{0.024} &0.028 &0.137 &0.136 &0.129 &0.143 &2.866 &4.018 &0.125 \\
ts225 &\textbf{0.012} &0.036 &0.043 &0.044 &0.034 &0.044 &0.884 &3.955 &0.036 \\
tsp225 &\textbf{0.019} &0.034 &0.041 &0.038 &0.033 &0.052 &1.036 &4.239 &0.031 \\
pr226 &\textbf{0.018} &0.031 &0.055 &0.052 &0.051 &0.056 &1.111 &1.102 &0.054 \\
gr229 &\textbf{0.019} &0.032 &0.091 &0.083 &0.088 &0.100 &1.996 &4.220 &0.080 \\
gil262 &\textbf{0.017} &0.043 &0.058 &0.061 &0.042 &0.068 &1.354 &4.640 &0.060 \\
pr264 &\textbf{0.030} &0.037 &0.066 &0.067 &0.053 &0.070 &1.401 &3.815 &0.055 \\
a280 &\textbf{0.032} &0.044 &0.524 &0.502 &0.473 &0.503 &10.064 &5.368 &0.471 \\
pr299 &\textbf{0.037} &0.062 &0.073 &0.060 &0.059 &0.088 &1.770 &5.429 &0.064 \\
lin318 &\textbf{0.026} &0.069 &0.091 &0.084 &0.079 &0.099 &1.977 &4.871 &0.090 \\
rd400 &\textbf{0.042} &0.096 &0.127 &0.112 &0.119 &0.151 &3.013 &7.594 &0.129 \\
fl417 &\textbf{0.048} &0.108 &0.211 &0.190 &0.192 &0.221 &4.425 &7.020 &0.193 \\
gr431 &\textbf{0.076} &0.133 &0.359 &0.353 &0.352 &0.396 &7.927 &8.016 &0.342 \\
pr439 &\textbf{0.094} &0.130 &0.231 &0.199 &0.194 &0.241 &4.817 &7.093 &0.197 \\
pcb442 &\textbf{0.088} &0.147 &0.193 &0.179 &0.151 &0.203 &4.055 &8.526 &0.157 \\
d493 &\textbf{0.144} &0.210 &0.884 &0.898 &0.886 &0.907 &18.141 &10.235 &0.889 \\
\textbf{att532} &\textbf{0.111} &0.219 &0.309 &0.265 &0.290 &0.329 &6.571 &10.010 &0.230 \\
u574 &\textbf{0.154} &0.218 &0.297 &0.272 &0.281 &0.339 &6.789 &10.652 &0.291 \\
rat575 &\textbf{0.105} &0.225 &0.236 &0.223 &0.186 &0.282 &5.647 &11.079 &0.204 \\
d657 &\textbf{0.159} &0.294 &1.025 &1.034 &1.003 &1.081 &21.617 &12.054 &1.028 \\
gr666 &\textbf{0.170} &0.381 &0.768 &0.651 &0.706 &0.808 &16.150 &12.513 &0.680 \\
u724 &\textbf{0.149} &0.394 &0.338 &0.359 &0.310 &0.465 &9.298 &12.734 &0.290 \\
rat783 &0.303 &0.448 &0.383 &0.389 &0.314 &0.532 &10.647 &14.551 &\textbf{0.299} \\
pr1002 &\textbf{0.498} &0.989 &0.879 &0.758 &0.766 &1.099 &21.974 &20.139 &0.747 \\
u1060 &\textbf{0.294} &0.802 &1.401 &1.198 &1.057 &1.444 &28.873 &20.074 &1.149 \\
vm1084 &\textbf{0.412} &0.887 &1.061 &0.867 &0.712 &1.152 &23.045 &17.057 &0.722 \\
pcb1173 &\textbf{0.402} &0.956 &1.109 &1.050 &0.997 &1.458 &29.156 &22.952 &0.947 \\
d1291 &\textbf{0.918} &1.247 &4.518 &4.452 &4.138 &4.289 &85.775 &20.517 &3.881 \\
rl1304 &\textbf{0.538} &1.303 &1.284 &1.145 &1.069 &1.574 &31.470 &19.792 &1.075 \\
rl1323 &\textbf{0.696} &1.332 &1.739 &1.655 &1.268 &1.861 &37.213 &20.215 &1.324 \\
nrw1379 &\textbf{0.994} &1.624 &1.270 &1.203 &1.107 &1.762 &35.249 &29.439 &1.113 \\
fl1400 &\textbf{0.745} &1.228 &2.655 &2.674 &2.559 &2.816 &56.325 &30.401 &2.583 \\
u1432 &\textbf{1.050} &1.917 &1.954 &1.760 &1.422 &2.104 &42.074 &32.086 &1.072 \\
fl1577 &\textbf{1.044} &2.117 &2.650 &2.616 &1.840 &2.713 &54.267 &25.673 &1.438 \\
d1655 &\textbf{1.551} &2.689 &8.131 &7.945 &7.751 &8.623 &172.450 &30.257 &7.624 \\
vm1748 &\textbf{0.660} &2.360 &2.525 &1.836 &1.914 &2.894 &57.884 &28.875 &1.967 \\
\noalign{\smallskip}
\hline
\noalign{\smallskip}
\textbf{average} &\textbf{0.219} &\textbf{0.428} &\textbf{0.708} &\textbf{0.665} &\textbf{0.612} &\textbf{0.769} &\textbf{15.374} &\textbf{9.822} &\textbf{0.594} \\
\textbf{std} &\textbf{0.351} &\textbf{0.679} &\textbf{1.364} &\textbf{1.319} &\textbf{1.251} &\textbf{1.433} &\textbf{28.660} &\textbf{9.280} &\textbf{1.217} \\
\bottomrule
\end{tabular}
\end{table}

To check the behaviour of the heuristics in relation to time, Table \ref{tab:time} shows the CPU time for each policy and heuristic shown in Table \ref{tab:1}. 
Note that for each query to the ResNet the input procedure produces an image that increases the computational burden.
Therefore, future work could be proposed to speed the ML component, even though the computation times remain short and acceptable for many online optimization scenarios.

Finally, to make a comparison with the metrics presented for the MF and CW in Section \ref{subsec:2.2}, 
results in Table 5 show the final tour achievements for the F, ML-C and ML-SC policies across the various positions in the CL.
Note that although the TPR of ML-SC is $100\%$, its FPR is not equal to zero since during the second phase some edges in first and second position can be inserted.
Also note that the accuracy of ML-C is consistently better than F, while the FPR for the first position is lower resulting in a higher TPR for the second position.

\begin{table}[!t]\centering
\caption{True positive rate (TPR), false positive rate (FPR), Accuracy (Acc) and positive likelihood rate (PLR) comparison across several positions and policies.}\label{tab:metrics ML-C}
\begin{tabular}{|c|c|cc|cc|}\specialrule{.1em}{.0em}{.0em}
position &method &TPR &FPR & Acc & PLR \\\specialrule{.1em}{.0em}{.0em}
\multirow{3}{*}{1} & F & 98.07\% & 85.80\% & 86.68\% & 1.1430 \\
&ML-C & 93.64\% & 53.08\% & 87.29\% &1.764 \\
&ML-SC & 100\%  & 3.41\%  & 99.53\% & 29.326 \\\hline
\multirow{3}{*}{2} & F & 68.28\% & 18.12\% & 73.62\% & 3.768 \\
&ML-C & 84.74\% & 29.07\% & 79.31\% &2.915 \\
&ML-SC & 100\% & 3.01\% & 98.82\% & 33.223 \\\hline
\multirow{3}{*}{3} & F & 44.40\% & 8.28\% & 80.39\% &5.362 \\
&ML-C & 42.74\% & 6.02\% &81.71\% &7.100 \\
&ML-SC & 86.27\% & 0.91\% & 96.02\% &94.802 \\\hline
\multirow{3}{*}{4} & F & 38.09\% & 5.75\% & 87.26\% &6.624 \\
&ML-C & 33.08\% & 3.60\% & 88.49\% &9.189 \\
&ML-SC & 80.27\% & 0.94\% & 96.726\% & 85.394 \\\hline
\multirow{3}{*}{5} & F & 31.18\% & 4.42\% & 91.95\% &7.054 \\
&ML-C & 26.52\% & 1.94\% & 94.03\% &13.670 \\
&ML-SC & 73.95\% & 0.88\% & 97.70\% &84.034 \\\hline
\multirow{3}{*}{>5} & F & 28.94\% & 0.02\% & 99.97\% &1447 \\
&ML-C & 25.76\% & 0.02\% &99.97\%  &1288 \\
&ML-SC & 64.24\% & 0.01\% & 99.99\% & 6424 \\\hline
\specialrule{.1em}{.0em}{.0em}
\end{tabular}
\end{table}

\section{Discussion}\label{sec:5}
A new strategy to design constructive heuristics has been presented. 
It gives a central role to the integration of statistical, mathematical and heuristic exploration.
We introduced a new way of thinking about the generalization of machine learning approaches for the TSP,
leading to an efficient integration between learning useful information and exploiting through classic approaches.
The objective is to learn useful skills from past experience to enhance the heuristic search. 
Our \emph{ML-Constructive} is the first machine learning approach able to scale and show improvements at once with respect to a classic efficient constructive heuristic.
Furthermore, the introduced approach is able to give good guidelines about how the machine learning can behave in the event of extreme negative or positive cases.
Results are very promising, and suggest that giving
more emphasis on the generalization of hybrid designs pays off.

The relevance of an exploratory stage with  statistical studies of the problem at hand had been emphasised.
The target of these studies is to select an effective sub-problem which allows the avoidance of many known machine learning flaws.

More work needs to be done to improve the accuracy and the extrapolation of the machine learning classifier.
Further improvements in future work could be in the direction of reducing the (constant) time required to prepare the input for the machine learning classifier, and to find the integration to meta-heuristics approaches as well.

\section*{Acknowledgement}
Umberto Junior Mele was supported by the Swiss National Science Foundation
through grants 200020-182360: ``Machine learning and sampling-based metaheuristics for stochastic vehicle routing problems''.

\appendix

\section{Complexity of the inner-loop constraint tracker}\label{complexity}


The purpose here is to compute the complexity of the inner-loop constraint tracker used in the \emph{ML-Constructive} heuristic process, as stated in Section \ref{subsec:2.4} and in Algorithm \ref{ML-Constructive}. 
For comprehension purposes we strict our analysis to the symmetric TSP, but similar results can be achieved for the asymmetric case as well.

\begin{figure}[!b]
    \centering
    \includegraphics[scale=0.3]{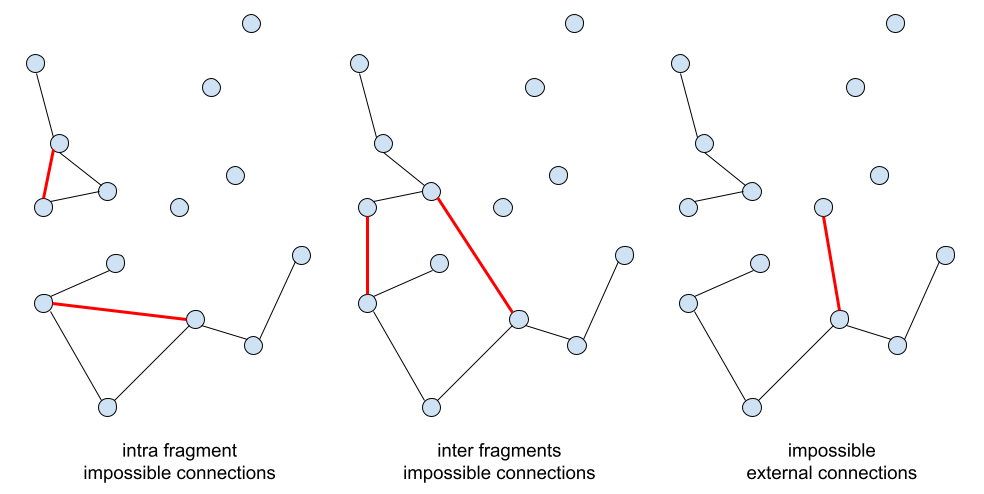}
    \caption{Events that cannot occur as an input to the tracking procedure}
    \label{fig:impossible cases}
\end{figure}

\begin{figure}[!t]
    \centering
    \includegraphics[scale=0.29]{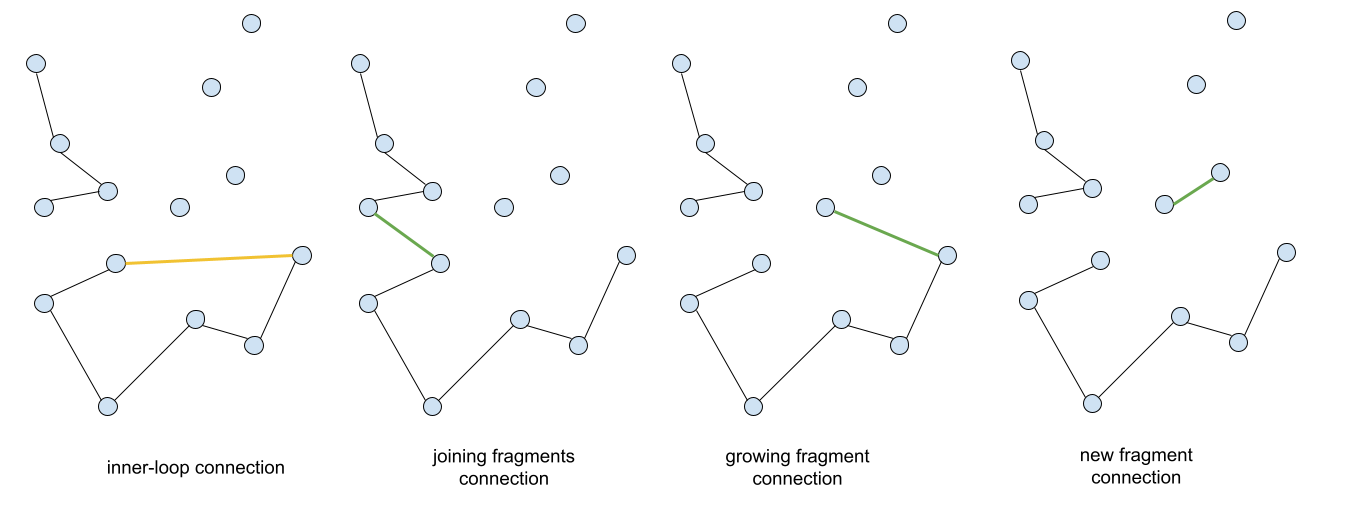}
    \caption{Events that can occur and are prevented by the tracking procedure (first two), and events that can be detected in constant time (last two).}
    \label{fig:possible cases}
\end{figure}

Firstly, we observe that the constraint tracker procedure is applied only to edges which have both extremes with exactly one connection already in the partial solution,
since the tracker routine follows the constraints expressed by lines $8$ and $22$ in Algorithm \ref{ML-Constructive}.
Therefore, edges connecting vertices from internal points of the fragments are impossible to occur at this point, as shown in Figure \ref{fig:impossible cases}.
While those creating an inner-loop and joining two fragments are possible to occur events (Figure \ref{fig:possible cases}).
Note that the goal of the tracker is to detect the inner-loop connections from the other.
The growing connections and new fragment connections shown in Figure \ref{fig:possible cases} are events that can be detected in constant time, since it's enough to check that an extreme of the inserting edge has zero connections in the partial solution (lines $12$ and $25$).
Also note that these two events cannot occur as input of the tracker procedure since they do not satisfy the constraint expressed by lines $8$ and $22$.

Secondly, we notice that the complexity of the worst case scenario for the whole procedure (from empty solution to the complete) is being computed in this Appendix.
Therefore, we are not taking in consideration just the single call of the tracker, but the global computation during the complete tracking process.
In fact, considering that the maximum number of positive addition for a constructive heuristic that growths fragments is equal to $n$ (the length of the tour).
Where, an addition is positive if the edge being attached to the partial solution complies the TSP constraints in (1b-e) and the ML decision-taker agrees to add the considered edge in solution. 
We refer to the epoch between two positive additions as $t$, i.e. no edge is in solution at $t=0$, meanwhile exactly eight edges are in solution at $t=8$. 
Take into consideration that the epochs to be checked by the tracking routine for the symmetric TSP are from $t=2$ to $t=n-2$.

\begin{figure}[!b]
    \centering
    \includegraphics[scale=0.3]{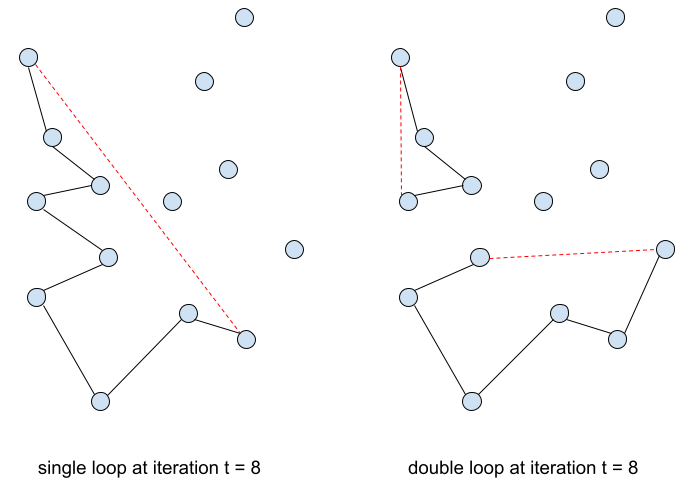}
    \caption{Single and double fragments possible inner-loops at the eighth epoch.}
    \label{fig:single vs double}
\end{figure}

As mentioned, the computationally expensive events that the tracker needs to check are the \enquote{inner-loop connection} and the \enquote{joining fragments connections}. 
The inner-loop connection drawn in yellow (Figure \ref{fig:possible cases}) occurs when the extremes of a fragment are connected together by the attaching edge $l$.
If we assume that at the epoch $t$ there are at most $s \leq t$ fragments, then there exist at most $s$ attaching edges at this epoch that can create an inner loop (Figure \ref{fig:single vs double}), and the sum of the operation needed to check these $s$ inner-loops is at most equal to $t$.
In fact, the tracker checks by spanning completely one of the fragment connecting to the attaching edge.
Then if the other extreme of the fragment coincide with the other extreme of the attaching edge there is an "inner loop connection", otherwise is a \enquote{joining fragments connection}.
Note that once an edge has been rejected  the fragment associated to it is set free for the current epoch, and the tracker do not need to check anymore its extremes. 
Since we have at most $t$ operations for epoch, and we have at most $n$ epochs, the global computation is $O(n^2)$.

Once the upper bound of the complexity for detecting the \enquote{inner-loop connections} has been found, the number of operations required for the \enquote{joining fragments connections} occurrences is still necessary to be estimated.
Usually, after having encountered a \enquote{joining fragments connection} event, the insertion of the considered edge $l$ takes place.
But since in \emph{ML-Constructive} it could happen that the ML decision-taker rejects the attaching edge (line 10), it may happen that the tracker is called many times during the same epoch.
Which could be a problem if the promising list $L_P$ was not limited at most $m \times n$ edges (Section \ref{subsec:2.3}).
Assuming that the worst case scenario is $O(n)$ for each edge processed in the first phase, we are still safe with $O(n^2)$ operations for the global tracking computation.

\section{The earlier insertion of the most promising edges could increase the probability of finding the optimal tour.}\label{first_probable}

The purpose of this Appendix is to present some advantages that a procedure which inserts promising edges into solution first have with respect to others approaches.
If the growing fragments heuristic is considered as a stochastic process, then we could estimate the probability that the optimal tour has of occurring following the procedure.
In fact, for each edge $l$ considered to be included in the partial solution there are two possible events: included or not.
If the random variable $E_l$ is used to refer to the event that the edge $l$ is included in solution ($\neg E_l$ otherwise), then we can express the probability that such event occurs as:

\begin{equation}\label{probE}
    P(E_l) = 1 - P(A_l) - P(B_l)  \quad \textrm{and} \quad  P(\neg E_l) = P(A_l) + P(B_l)
\end{equation}

\noindent where $A_l$ refers to the internal point connection events  (Figure \ref{fig:impossible cases}), while $B_l$ stand for the inner-loop connection events (Figure \ref{fig:possible cases}).
Recall that internal point connection occurs when the constraint which ensures that no vertex is connected to more than two other vertices is not satisfied.
While an inner-loop occurs when sub-solutions are created, instead of having a single global loop.

In case that a list $L$ is used to store all the existing edges of the TSP instance that we wish to solve, and we randomly shuffle such list to create a random examination order.
The probabilities of the events $A_l$ and $B_l$ will be dependent on the position $p$ in such list and the number of edges already inserted in the partial solution.
So, combinatorics can help us calculate or approximate these probabilities.
Recalling the epoch concept described in Appendix \ref{complexity}, we can state that at $t=0$ such probability is one, while at $t=n$ the probability of $E_l$ is null:

\begin{equation}
    P(E_l | \, t=0) = 1
\end{equation}
\begin{equation}
    P(E_l | \, t=n) = 0
\end{equation}

\noindent Since at $t=0$ no edge has been placed in solution, no $A_l$ or $B_l$ event can occur. 
While at t=n the solution is complete.
Then, we want to prove that the probability of $E_l$ will monotonically decrease as more edges are fed into the solution and as we progress through the $L$ list.
In case this conjecture is true, we can conclude that edges inspected earlier in the list are more likely to be included than those seen later.
Emphasising the need to put first in the $L$ list the edges that we consider most promising to be included in the optimal solution.

As a first step, we determine the probability of occurrence of $A_l$.
To figure out such probability, we shall simply estimate the number of cases in which $A_l$ occurs and divide by the total number of possible cases.
These cases vary depending on the position $p$ in the list, the number of edges $e$ and the number of vertices $n$ in the instance. 
Recalling that $A_l$ occurs when in the list the edge $l$ is preceded by at least $2$ other edges that have the same extreme with that of $l$.
In case there are $d$ of these overlapping edges, we have that $A_l$ occur for $d=2$ to $d=n-1$:

\begin{equation}
    P(A_l| \, p ) = \frac{\sum_{d=2}^{n-1} \binom{n-1}{d} \binom{e-n +1}{p- 1 -d} }{\binom{e}{p-1}} \approx \sum_{d=2}^{n-1} \frac{(p-1)!}{(p- 1 - d)!}
\end{equation}


\noindent which is an increasing function with respect to the position $p$, and converge to $1$ as $p$ goes to $e$. 

Meanwhile, to compute the probability of $B_l$, the epoch in which the event occurs must be taken into account. 
Bearing in mind that as we proceed along the p positions in the list such epoch is ascending, since there are no operations that remove an edge from the solution and it is possible just to add a new edge into such partial solution.

\noindent Considering that the maximum total number of inner-loops for a given epoch is fixed and equal to t, we have that:

\begin{equation}
    P(B_l | \, p, \, t) < \frac{t}{e - p -1}  \qquad \textrm{with}  \quad t \leq n
\end{equation}

\noindent which has upper bound that is an increasing function with respect to the position $p$ and the epoch $t$. 

To conclude, since the probabilities of $A_l$ and $B_l$ show an increasing trend, although not strictly due to the upper bound of $B_l$,
we can verify that the probability of $E_l$ has a decreasing trend due to Equation \ref{probE}.
Therefore, the anticipation of the insertion of promising edges is a good strategy for the heuristic.
However, such results do not prove that for any solving algorithm, the probability $P(E_L | \, t)$ is a strict decreasing function.
But it suggests that a general decreasing trend is present which should be exploited by the \emph{ML-Constructive} heuristic during the construction of the solution.

\end{document}